\documentclass[runningheads]{llncs}
\usepackage[T1]{fontenc}
\usepackage{booktabs}
\usepackage{multirow}
\usepackage{mathtools}
\usepackage{colortbl}
\usepackage{amssymb}

\usepackage[pagebackref,breaklinks,colorlinks]{hyperref}
\usepackage[dvipdfmx]{graphicx}

\usepackage{orcidlink}

\begin{document}
\newcommand{\repeatthanks}{\textsuperscript{\thefootnote}}
\title{DeBiFormer: Vision Transformer with Deformable Agent Bi-level Routing Attention}

\author{
NguyenHuu BaoLong\thanks{Equal contribution.} \inst{1} \and
Chenyu Zhang\repeatthanks \inst{1}\and
Yuzhi Shi \inst{1} \and 
Tsubasa Hirakawa \inst{1} \and
Takayoshi Yamashita \inst{1} \and 
Tohgoroh Matsui \inst{1} \and
Hironobu Fujiyoshi \inst{1}
}

\authorrunning{NguyenHuu B. et al.}
\titlerunning{DeBiFormer}
\institute{Chubu University, Kasugai, Japan \\
\email{maclong01@gmail.com} \\
\email{\{zhang1121, shi, hirakawa\}@mprg.cs.chubu.ac.jp} \\
\email{\{takayoshi, matsui, fujiyoshi\}@isc.chubu.ac.jp}}

\maketitle
\begin{abstract}
Vision Transformers with various attention modules have demonstrated superior performance on vision tasks. While using sparsity-adaptive attention, such as in DAT, has yielded strong results in image classification, the key-value pairs selected by deformable points lack semantic relevance when fine-tuning for semantic segmentation tasks. The query-aware sparsity attention in BiFormer seeks to focus each query on top-k routed regions. However, during attention calculation, the selected key-value pairs are influenced by too many irrelevant queries, reducing attention on the more important ones.
To address these issues, we propose the Deformable Bi-level Routing Attention (DBRA) module, which optimizes the selection of key-value pairs using agent queries and enhances the interpretability of queries in attention maps. Based on this, we introduce the Deformable Bi-level Routing Attention Transformer (DeBiFormer), a novel general-purpose vision transformer built with the DBRA module. DeBiFormer has been validated on various computer vision tasks, including image classification, object detection, and semantic segmentation, providing strong evidence of its effectiveness.Code is available at \href{https://github.com/maclong01/DeBiFormer}{https://github.com/maclong01/DeBiFormer}

% Vision Transformers with various attention modules have shown superior performances on vision tasks. Although using sparsity adaptive attention e.g., in DAT, has achieved great performance in image classification, the key and value pairs which selected by deformable points, are not semantical when fine-tuning in semantic segmentation tasks. The query-aware sparsity attention adopted in BiFormer aims for each query to focus on the key-value pairs based on top-k routed regions. However, when calculating the attention, the selected key-value pairs are influenced by too many less relevant queries, which reduces the attention allocated to important queries.
% To mitigate these problems, we propose the Deformable Bi-level Routing Attention (DBRA) module that could optimize the selection of key and value pairs using agent queries and enhance the interpretability of queries in attention maps. On this basis, a novel general-purpose vision transformer called \textbf{Deformable Bi-level Routing Attention Transformer} (DeBiFormer), built by the proposed DBRA, is introduced. Our DeBiFormer is validated across a range of computer vision tasks, including image classiﬁcation, object detection, and semantic segmentation. These results provide substantial evidence of its effectiveness. Code is available at \href{https://github.com/maclong01/DeBiFormer}{https://github.com/maclong01/DeBiFormer}
\end{abstract}

\section{Introduction}
The Vision Transformer has recently demonstrated significant promise in the realm of computer vision \cite{PVIT,SWIN,PVT}. It can capture long-range dependency in data \cite{SWIN,VIT}, and is almost leading to a convolution-free model more flexible for fitting tons of data \cite{PVT}. In addition, it enjoys high parallelism, which benefits training and inference for large models \cite{BERT,VIT}. 
The computer vision community has observed a surge in the adoption and development of Vision Transformers \cite{E2E,CSWIN,PVIT,SWIN,PVT,CROSS}.
%%%%%%%%%%%%%%%%%%%%%%%%%%%%%%%%%%%%%%%%%%%%%%%%%%%%%%%%%%%%%%%%%%%%%%%%%

\label{sec:intro}
\begin{figure*}[tb]
\begin{center}
\includegraphics[width=0.8\linewidth]{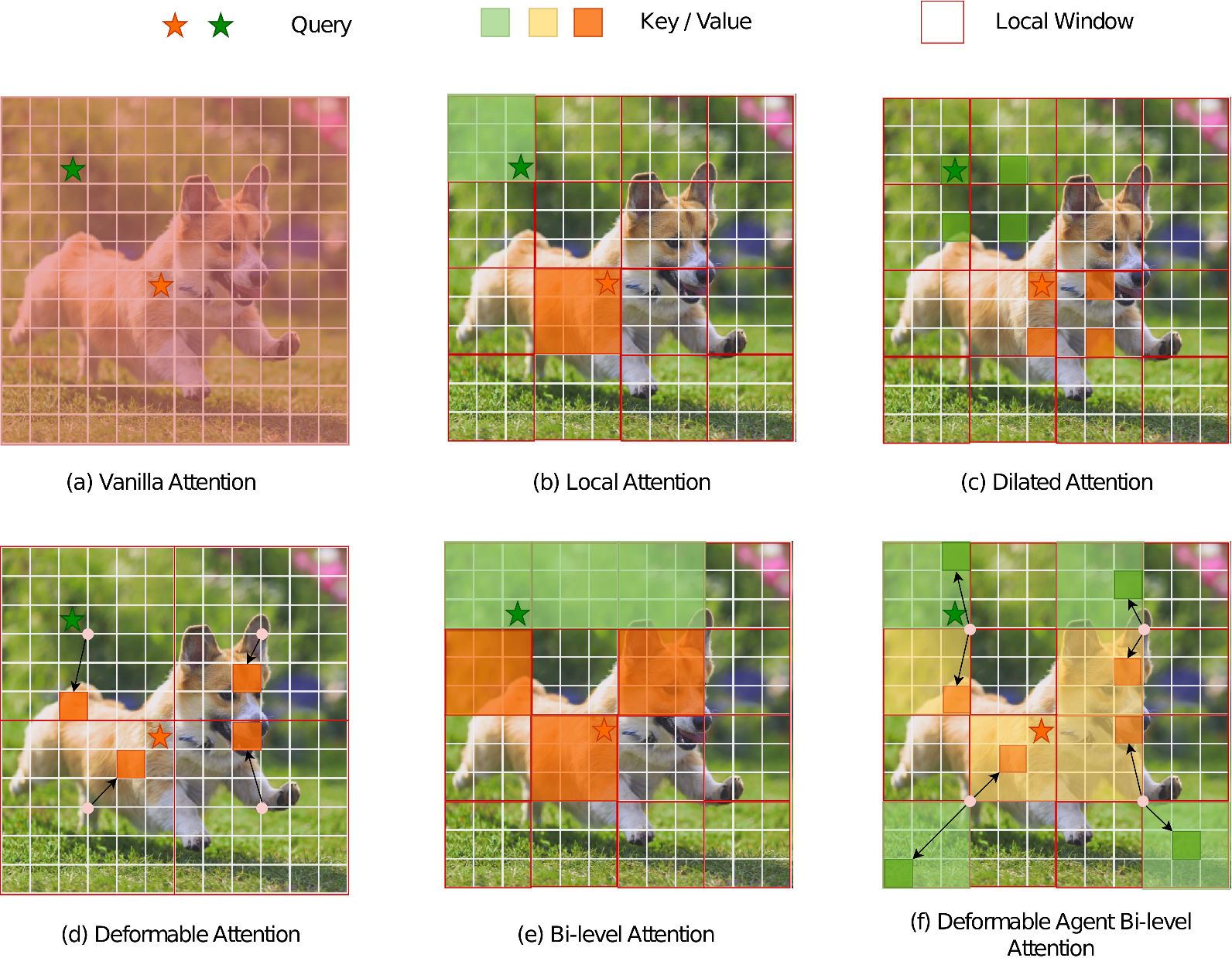}
\end{center}
   \caption{Vanilla attention and its sparse variants are illustrated in the diagram: (a) vanilla attention functions globally, leading to increased computational complexity and substantial memory footprint. (b)-(c) Multiple strategies strive to reduce complexity by incorporating sparse attention with various handcrafted patterns, such as local window \cite{LOCAL} and dilated window \cite{CROSS,MAXVIT,DILATEDFORMER}. (d) Deformable attention \cite{DEF} facilitates image-adaptive sparsity by deforming the regular grid. (e) Bi-level routing attention \cite{BIFORMER} begins by searching for top-k (k = 3 in this case) relevant regions and subsequently attends to the union of these regions. (f) In our approach, we realize bi-level routing attention, where the initial step involves searching for top-k (k = 1 in this case) relevant regions. Subsequently, attention is directed to the union of these regions by deforming regular grid attendance via top-k relevant regions.}
\label{attentionmethod}
\end{figure*}
To improve attention, numerous pieces of research have used thoughtfully crafted, efficient attention patterns in which each query selectively focused by a smaller portion of key-value pairs. As shown in Figure\ref{attentionmethod}, among the various representation approaches, some include local windows \cite{LOCAL} and dilated windows \cite{CROSS,MAXVIT,DILATEDFORMER}. In addition, some research has taken a different path through sparsity adaptation to data in their methodology, as demonstrated in the works of \cite{DEFPATCH,DEF}. However, despite the varying strategies for merging or selecting key and value tokens. These tokens are not semantic for queries. With this approach, when applied to other downstream tasks for pretrained ViT \cite{VIT} and DETR \cite{E2E}. Queries do not originate from semantic-region key-value pairs. Consequently, compelling all queries to focus on insufficient sets of tokens may not yield the most optimal results. Recently, with the dynamic query-aware sparsity attention mechanism, queries are focused by the most dynamic semantically key-value pairs, which is referred to as bi-level routing attention \cite{BIFORMER}. 
However, in this approach, queries are handled by semantic key-value pairs instead of originating from detailed regions, which may not yield the most optimal results in all cases. In addition, when calculating the attention, these keys and values selected for all queries are influenced by too many less relevant queries, resulting in a decrease of attention for important queries, which has a significant impact when performing segmentations.\cite{biunet,lan2024brau}.
 
To make the attention for queries more efficient, we propose the Deformable Bi-level Routing Attention (DBRA), an attention-in-attention architecture for visual recognition. During the process in DBRA, the first problem is how to locate deformable points. We use the observation in \cite{DEF} that attention has an offset network that takes as input query features and generates corresponding offsets for all reference points. Thus, candidate deformable points are shifted towards important regions with high flexibility and efficiency to capture more informative features. 
%%%%%%%%%%%%%%%%%%%%%%%%%%%%%%%%%%%%%%%%%%%%%%%%%%%%%%%%%%%%%%%%%%%%%%%%%%%%%%%%%%%%%%%%%%%%%%%%%%%%%%%%%%
The second problem is how to aggregate information from semantically relevant key-value pairs, and then broadcast the information back to queries.
%The second problem, as shown above, is that deformable attention needs to be enhanced for the segmentation downstream task. 
%Therefore, we propose an attention-in-attention architecture in which the subset queries are tokens that are shifted toward deformable points as shown above. 
Therefore, we propose an attention-in-attention architecture that shifted toward deformable points as shown above acts as the agent for the queries.
%%%%%%%%%%%%%%%%%%%%%%%%%%%%%%%%%%%%%%%%%%%%%%%%%%%%%%%%%%%%%%%%%%%%%%%%%%%%%%%%%%%%%%%%%%%%%%%%%%%%%%%%%%
As the key-value pairs are selected for deformable points, we use the observation in \cite{BIFORMER} to select a small portion of the most semantically relevant key-value pairs that a region only needs by focusing on the top-k routed regions. 
Then, with the semantically relevant key-value pairs selected, we first apply a token-to-token attention with deformable points queries. 
And then, we apply a second token-to-token attention to broadcast the information back to queries, in which deformable points as key-value pairs are designed to represent the most important points in a portion of semantic regions.
%we propose an attention method that integrates a deformable attention \cite{DEF} method with a bi-level routing \cite{BIFORMER} attention method in 
%which queries are attended to by the essential deformable points via sampling offsets to maintain performance in classification. Then, to provide semantics for deformable points, we focus on deformable points for other semantically relevant key-value pairs to increase the semantics. which deformable points are extracted from queries via sampling offsets and act as the agent for the query tokens to aggregate information from semantically relevant key-value pairs, then broadcast the information back to queries.
%%%%%%%%%%%%%%%%%%%%%%%%%%%%%%%%%%%%%%%%%%%%%%%%%%%%%%%%%%%%%%%%%%%%%%%%%%%%%%%%%%%%%%%%%%%%%%%%%%%%%%%%%%
%-------------------

To summarize, our contributions are as follows: 

1. We propose Deformable Bi-level Routing Attention (DBRA), an attention-in-attention architecture for visual recognition, where data-dependent attention patterns are obtained flexibly and semantically.

2. By utilizing the DBRA module, we propose a novel backbone, called DeBiFormer, which has a stronger recognition ability based on the visualization results of the attention heat map.

3. Extensive experiments on ImageNet \cite{IMAGENET}, ADE20K \cite{ADE20K}, and COCO \cite{COCO} demonstrate that our model consistently outperforms other competitive baselines. 

%including the Swin-T by a margin of $2.6\%$ in terms of top-1 accuracy for image classification, $5.5\%$ on the mIoU for semantic segmentation, and $4.4\%$ on object detection for both box AP and mask AP. The advantages on small and large objects are more distinct with a margin of $5.1\%$, $6.2\%$.
%-------------------------------------------------------------------------
%-------------------------------------------------------------------------
\section{Related Work}
\subsection{Vision Transformers} The Transformer-based backbone incorporates channel-wise MLP \cite{MLPMIX} blocks to embed per-location features through channel mixing. Additionally, attention \cite{VIT} blocks are used for cross-location relation modeling and facilitating spatial mixing. Initially devised for natural language processing \cite{VIT,BERT}, Transformers were subsequently introduced to the domain of computer vision through works like DETR \cite{E2E} and ViT \cite{VIT}. Compared with CNNs, the primary distinction lies in the fact that transformers use attention as a substitute for convolution, thereby facilitating global context modeling. Nevertheless, vanilla attention, which calculates pairwise feature affinity across all spatial locations, imposes a significant computational burden and leads to substantial memory footprints, particularly when dealing with high-resolution inputs. Thus, a key research focus is to devise more efficient attention mechanisms, crucial for mitigating computational demands, especially with high-resolution inputs.

\subsection{Attention mechanisms} Numerous studies have aimed to alleviate the computational and memory complexities associated with vanilla attention. Approaches include sparse connection patterns \cite{CONNECTPATTERN}, low-rank approximations \cite{LINFORMER}, and recurrent operations \cite{TXL}. In the context of Vision Transformers, sparse attention has gained popularity, particularly following the remarkable success of the Swin Transformer \cite{SWIN}. Within the Swin Transformer framework, attention is constrained to non-overlapping local windows, and an innovative shift window operation is introduced. This operation facilitates communication between adjacent windows, contributing to its unique approach to handling attention mechanisms. To attain larger or approximate global receptive fields without exceeding computational constraints, recent studies have incorporated diverse manually designed sparse patterns. These include the integration of dilated windows \cite{CROSS,MAXVIT,DILATEDFORMER} and cross-shaped windows \cite{CSWIN}. Moreover, certain studies endeavor to make sparse patterns adaptable to data, as demonstrated by works like DAT \cite{DEF}, TCFormer \cite{TCFORMER}, and DPT \cite{DEFPATCH}. Despite their efforts to decrease the number of key-value tokens using diverse merging or selection strategies, it is crucial to recognize that these tokens lack semantic specificity. Instead, we reinforce query-aware key-value token selection. 

Our work is motivated by an observation: semantically attentive regions for important queries can exhibit significant differences, as illustrated by visualizations from pre-trained models like ViT \cite{VIT} and DETR \cite{E2E}. In achieving query-adaptive sparsity through a coarse-to-fine approach, we propose an attention-in-attention architecture which utilizes the deformable attention \cite{DEF} with bi-level routing attention \cite{BIFORMER}. Diverging from deformable attention \cite{DEF} and bi-level routing attention \cite{BIFORMER}, our deformable bi-level routing attention aims to reinforce the most semantic and flexible key-value pairs. In contrast, bi-level routing attention only focuses on locating a few highly relevant key-value pairs, while deformable attention prioritizes identifying a few of the most flexible key-value pairs. 

\section{Our Approach: DeBiFormer}
%%%%%%%%%%%%%%%%%%%%%%%%%%%%%%%%%%%%%%%%%%%%%%%%%%%%%%%%%%%%%%%%%%%%%%%%%
\begin{figure*}[!ht]
\begin{center}
\includegraphics[width=0.9\linewidth ]{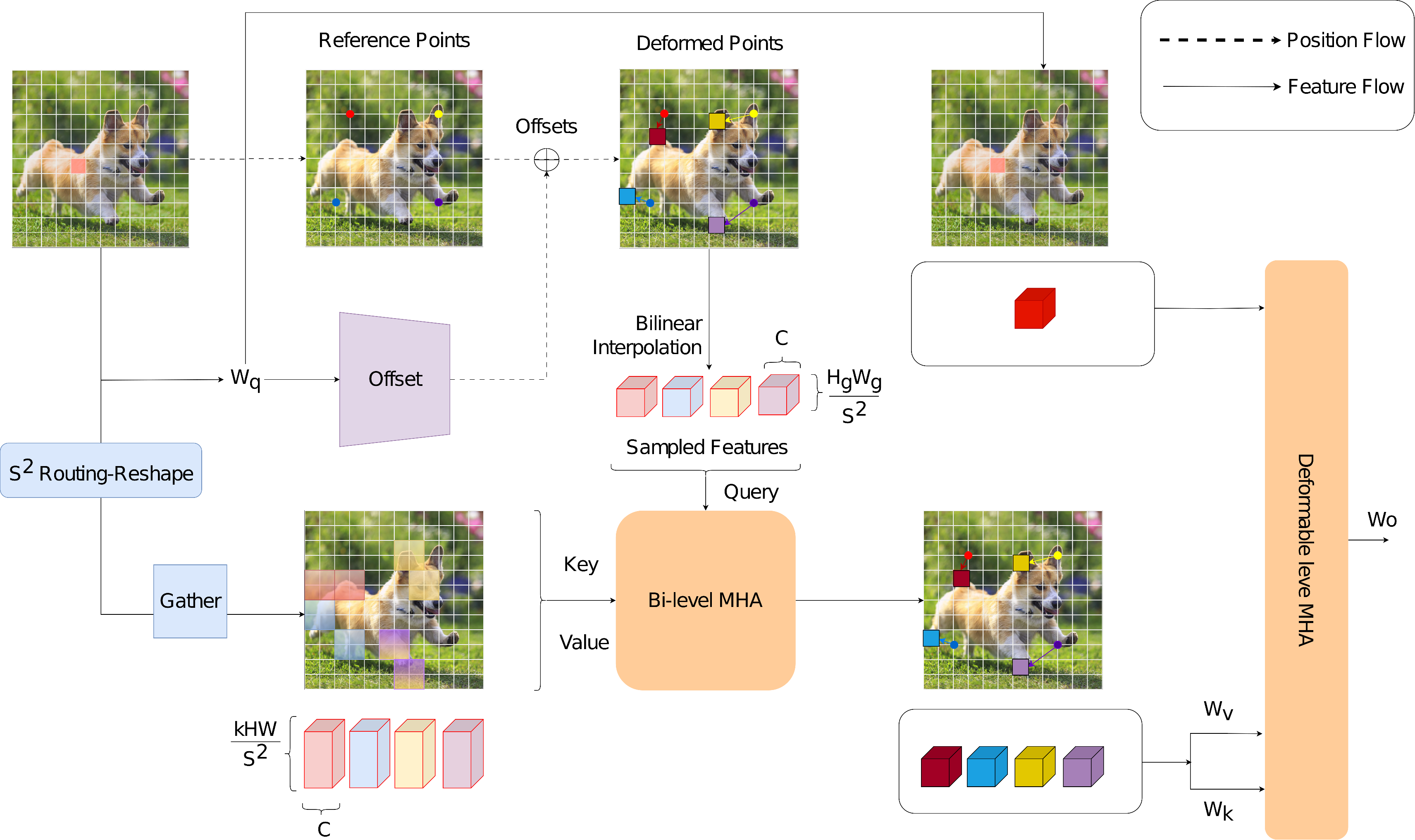}
\end{center}
   \caption{Detailed architecture of Deformable Bi-level Routing Attention. In the top-left part, the set of reference points is uniformly distributed across the feature map. Offsets for these points are learned from queries through the offset network. Then, in the top-middle part, deformed features are projected from sampled features based on the locations of deformed points. In the bottom-left-middle part, we attend to projected deformed features by utilizing gathered key-value pairs in top-k-related windows. 
}
\label{debi-module}
\end{figure*}
%%%%%%%%%%%%%%%%%%%%%%%%%%%%%%%%%%%%%%%%%%%%%%%%%%%%%%%%%%%%%%%%%%%%%%%%% 
\subsection{Preliminaries}
Initially, we revisit the attention mechanism used in recent Vision Transformers. Taking a flattened feature map $x \in \mathrm{R}^{N \times C}$ as the input, a multi-head self-attention (MHSA) block with $M$ heads is formulated as
%%%%%%%%%%%%%%%%%%%%%%%%%%%%%%%%%%%%%%%%%%%%%%%%%%%%%%%%%%%
\begin{gather}
q = xW_{q}, k = xW_{k}, v=xW_{v},\\
z^{(m)} = \sigma(q^{(m) k^{(m)\top}/\sqrt{d}})v^{(m)}, m=1,...,M,\\
z = Concat(z^{(1)},...,z^{(M)})W_{o}
\label{eq3}
\end{gather}
%%%%%%%%%%%%%%%%%%%%%%%%%%%%%%%%%%%%%%%%%%%%%%%%%%%%%%%%%%%
where $\sigma(\cdot)$ denotes the softmax function, and $d = C/M$ is the dimension of each head. $z(m)$ denotes the embedding output from the $m$-th attention head, and $q(m), k(m), v(m) \in \mathrm{R}^{N\times d}$ denote the query, key, and value embeddings, respectively. $W_q, W_k, W_v, W_o \in \mathrm{R}^{C\times C}$ are the projection matrices. With normalization layers and identity shortcuts, the $l$-th Transformer block, for which LN means layer normalization, is formulated as
\begin{gather}
z'_{l} = MHSA(LN(z_{l-1})) + z_{l-1},\\
z_{l} = MLP(LN(z'_{l}))+z'_{l}
\end{gather}
%%%%%%%%%%%%%%%%%%%%%%%%%%%%%%%%%%%%%%%%%%%%%%%%%%%%%%%%%%%%%%%%%%%%%%%%%
%---------------------------------------------------------------------------------------------------------------------------
\subsection{Deformable bi-level routing attention (DBRA)}
The architecture of the proposed Deformable Bi-level Routing Attention (DBRA) is illustrated in Figure \ref{debi-module}. We first employ a deformable attention module, which includes an offset network that generates offsets for reference points based on the query features, creating deformable points. However, these points tend to cluster in important regions, leading to an over-concentration in certain areas.

To address this, we introduce deformable points-aware region partitioning, ensuring that each deformable point interacts with only a small subset of key-value pairs. Yet, solely relying on region partitioning can result in an imbalance between important and less important regions. To tackle this, the DBRA module is designed to distribute attention more effectively. In DBRA, each deformable point acts as an agent query, computing attention with semantic region key-value pairs. This approach ensures that only a few deformable points are assigned to each important region, allowing attention to be spread across all critical areas of the image rather than clustered in one spot.

By employing the DBRA module, attention is reduced in less important regions and increased in more important ones, ensuring a balanced distribution of attention throughout the image.
% The detailed architecture of the proposed Deformable Bi-level Routing Attention(DBRA) is shown in Figure\ref{debi-module}. We first utilize a deformable attention module whose attention has an offset network that takes as input query features and generates corresponding offsets for all reference points to create deformable points.
% The shortcoming of using deformable points is that these points will cluster in an important region, making too many deformable points in one area. 
% Therefore, we conduct deformable points aware regions partition aiming for each deformable point focused by a small portion of key-value pairs.
% But only using region partition will lead to an imbalance between important and unimportant regions. Hence, we propose the DBRA module to deal with this problem. 
% In our DBRA module, one deformable point will act as an agent query, and calculate the attention with semantic region key-value pairs, which makes it possible for each important region only need few deformable points and these deformable points can be distributed to all important regions of the entire image, rather than clumping all the deformable points into a single important region.
% By using our DBRA module, which reduces attention to the unimportant regions and increases attention to the important regions, the selection of attention is distributed in a balanced manner.
\\
\textbf{Deformable attention module and input projection.} 
As illustrated in Fig.\ref{debi-module}, given the input feature map $x \in \mathrm{R}^{H\times W\times C}$, a uniform grid of points $p \in \mathrm{R}^{H_{G}\times W_{G}\times 2}$ is generated by downsampling the input feature map by factor $r$, $H_{G}=H/r, W_{G}=W/r$ as a reference. To obtain the offset for each reference point, the features are linearly projected to generate query tokens $q=xW_{q}$, which are then input into the $\theta_{offset}(\cdot)$ subnetwork to produce the offsets $\Delta p=\theta_{offset}(q)$. Subsequently, the features are sampled at the locations of deformed points as keys and values and further processed by projection matrices:
\begin{equation}
q=xW_{q}, \Delta p=\theta_{offset}(q), \bar{x}= \varphi(x;p+\Delta p),
\end{equation}
where $\bar{x}$ represent the deformed key $\bar{k}$ and value $\bar{v}$ embeddings, respectively. Specifically, we set the sampling function $\varphi(\cdot;\cdot)$ to a bilinear interpolation to make it differentiable:
\begin{equation}
\varphi(z;(p_{x},p_{y}))=\sum_{r_{x},r_{y}}g(p_{x},r_{x})g(p_{y},r_{y})z[r_{y},r_{x},:],
\label{eq8}
\end{equation}
where the function $g(a,b)=max(0,1-|a-b|)$ and $(r_{x},r_{y})$ represent indices for all locations on $z \in \mathrm{R}^{H\times W\times C}$. In a similar setup as deformable attention, where $g$ is nonzero on the four integral points closest to $(px, py)$, Equation \ref{eq8} is simplified to a weighted average across these four locations.\\
%%%%%%%%%%%%%%%%%%%%%%%%%%%%%%%%%%%%%%%%%%%%%%%%%%%%%%%%%%%%%%%%%%%%%%%%%
\textbf{Region partition and region-to-region routing.}
With the deformable attention feature map input $\bar{x}\in \mathrm{R}^{H_{G}\times W_{G}\times C}$ and feature map $x \in \mathrm{R}^{H \times W \times C}$, the process begins by dividing it into regions of size $S\times S$ non-overlapped regions such that each region contains $\frac{H_{G}W_{G}}{S^{2}}$ feature vectors with reshaped $\bar{x}$ as $\bar{x^{r}}\in \mathrm{R}^{S^{2}\times \frac{H_{G}W_{G}}{S^{2}}}\times C$ and $x$ as $x^{r}\in \mathrm{R}^{S^{2}\times \frac{HW}{S^{2}}}\times C$. Then, we derive the query, key, and value with linear projections:
%%%%%%%%%%%%%%%%%%%%%%%%%%%%%%%%%%%%%%%%%%%%%%%%%%%%%%%%%%%%%%%%%%%%%%%%%
\begin{equation}
\hat{q}=\bar{x^{r}}W_{q},\hat{k}=x^{r}W_{k},\hat{v}=x^{r}W_{v}.
\end{equation}
Next, we use the region-to-region method, as introduced in BiFormer \cite{BIFORMER}, which is applied to establish the attending relationship by constructing a directed graph. To initiate the process, region queries and keys $\hat{q}^{r},\hat{k}^{r}\in \mathrm{S}^{S^{2}\times C}$ are derived through the application of per-region averaging. Subsequently, the adjacency matrix $A^{r}\in \mathrm{S^{2}\times S^{2}}$ for the region-to-region affinity graph is derived through $Q^{r}$ and $K^{r^{\top}}$ matrix multiplication:
\begin{equation}
A^{r}=\hat{q}^{r}(\hat{k}^{r})^{\top},
\end{equation}
where adjacency matrix $A^{r}$ quantifies the semantic relationship between two regions. The crucial step in this method involves pruning the affinity graph by retaining only the $topk$ connections for each region with a routing index matrix $I^{r}\in \mathrm{N}^{S^{2}\times k}$ through the use of the $topk$ operator:
\begin{equation}
I^{r}=topk(A^{r}).
\end{equation}
%%%%%%%%%%%%%%%%%%%%%%%%%%%%%%%%%%%%%%%%%%%%%%%%%%%%%%%%%%%%%%%%%%%
\textbf{Bi-level token to deformable-level token attention.} 
Utilizing the region routing matrix $I^{r}$, we can then apply token attention. For each deformable query token within region $i$, its attention spans all key-value pairs located in the $topk$ routed regions, that is, those indexed by $I^{r}_{i,1},I^{r}_{i,2},...,I^{r}_{i,k}$. Hence, we continue the process of gathering the key and value:
\begin{equation}
\hat{k}^{g}=gather(\hat{k},I^{r}), \hat{v}^{g}=gather(\hat{v},I^{r}),
\end{equation}
where $\hat{k}^{g},\hat{v}^{g}\in \mathrm{R}^{S^{2}\times \frac{kHW}{S^{2}}\times C}$ are the gathered key and value. Then, we apply attention on $\hat{k}^{g},\hat{v}^{g}$ as:
\begin{gather}
\hat{O}=\hat{x}+W_{o'}(Attention(\hat{q},\hat{k}^{g},\hat{v}^{g})+LCE(\hat{v})),\\
O=MLP(LN(\hat{O}))+\hat{O},
\end{gather}
where $W_{o'}$ is a projection weight for output features, and $LCE(\cdot)$ uses a kernel size 5 depth-wise convolution. 
\\
\textbf{Deformable-level token to token attention.} 
Following that, the deformable features that are semantically attended to via \cite{BIFORMER} are reshaped $O$ as $O^{r}\in \mathbb{R}^{H_{G}\times W_{G}\times C}$ and parameterized at the locations of keys and values:
\begin{equation}
k=O^{r}W_{k},v=O^{r}W_{v},
\end{equation}
$k$ and $v$ represent the embeddings of semantically deformed keys and values, respectively. Using existing approaches, we perform self-attention on $q, k, v$, and relative position offsets $R$. The output of attention is formulated as follows:
\begin{equation}
z^{m}=W_{\bar{o}}(\sigma(q^{m}k^{(m)\top}/\sqrt{d}+\phi(\hat{B};R))v^{m}).
\end{equation}
Here, $\phi(\hat{B}; R)\in \mathbb{R}^{HW\times H_{G}W_{G}}$ corresponds to the position embedding, following the approach of previous work \cite{SWIN}. Then $z^{m}$ is projected though $W_{o}$ to get the final output $z$ as Equation \ref{eq3}.
%-------------------------------------------------------------------------
\subsection{Model architectures}
%%%%%%%%%%%%%%%%%%%%%%%%%%%%%%%%%%%%%%%%%%%%%%%%%%%%%%%%%%%%%%%%%%%%%%%%%%%%%%%%%%%%%%%%%%%%%%%%%%%%%%%%%%%%%%%%%%%%%%
\begin{figure}[!tb]
\begin{center}
\includegraphics[width=1.0\linewidth]{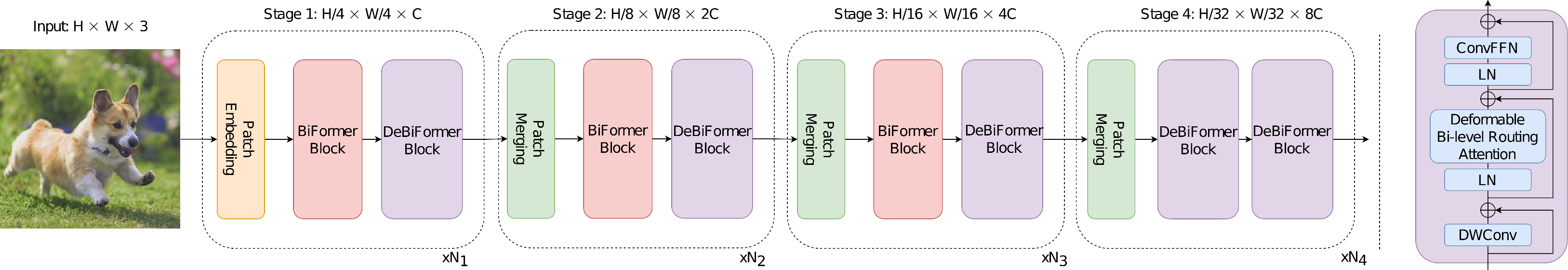}
\end{center}
   \caption{Overall model architecture of our DeBiFormer. \textbf{Left:} Network architecture of DeBiFormer. $N_{1}$ to $N_{4}$ represent numbers of stacked successive local and Deformable Bi-level Routing Attention blocks. Please consult Table \ref{table-architecture} for specific configurations. \textbf{Right:} Details on DeBiFormer Block.}
\label{debi-stage}
\end{figure}
%%%%%%%%%%%%%%%%%%%%%%%%%%%%%%%%%%%%%%%%%%%%%%%%%%%%%%%%%%%%%%%%%%%%%%%%%%%%%%%%%%%%%%%%%%%%%%%%%%%%%%%%%%%%%%%%%%%%%%
\begin{table}[!t]
\small
\centering
\begin{tabular}{cccc}
\hline
\multicolumn{4}{c}{Variant Architectures of DeBiFormer}                                                                                   \\
\multicolumn{1}{c|}{} & \multicolumn{1}{c|}{DeBi-T}                                                                           & \multicolumn{1}{c|}{DeBi-S}    & DeBi-B      \\ \hline
\multicolumn{1}{c|}{\multirow{3}{*}{\begin{tabular}[c]{@{}c@{}}Stage 1\\ 56$\times$56 \\ \end{tabular}}} 

& \multicolumn{1}{c|}{\multirow{3}{*}{\begin{tabular}[c]{@{}c@{}}$N_{1}$=2, C=64\\ r=8, M=2, $D_{r}$=3\\ G=1, K=9, $B_{r}$=3\end{tabular}}}   

& \multicolumn{1}{c|}{\multirow{3}{*}{\begin{tabular}[c]{@{}c@{}}$N_{1}$=4, C=64\\ r=8, M=2, $D_{r}$=3\\ G=1, K=9, $B_{r}$=3\end{tabular}}} 

& \multirow{3}{*}{\begin{tabular}[c]{@{}c@{}}$N_{1}$=4, C=96\\ r=8, M=3, $D_{r}$=3\\ G=1, K=9, $B_{r}$=3\end{tabular}}   \\

\multicolumn{1}{c|}{}                                                                                
& \multicolumn{1}{c|}{}                                                                                                    
& \multicolumn{1}{c|}{}                                                                                                    
& \\
\multicolumn{1}{c|}{}  
& \multicolumn{1}{c|}{}                                                                                                    
& \multicolumn{1}{c|}{}                                                                                                    
&                                                                                                     
\\ \hline
\multicolumn{1}{c|}{\multirow{3}{*}{\begin{tabular}[c]{@{}c@{}}Stage 2\\ 28$\times$28\\ \end{tabular}}} 
& \multicolumn{1}{c|}{\multirow{3}{*}{\begin{tabular}[c]{@{}c@{}}$N_{2}$=2, C=128\\ r=4, M=4, $D_{r}$=3\\ G=2, K=7, $B_{r}$=3\end{tabular}}}  

& \multicolumn{1}{c|}{\multirow{3}{*}{\begin{tabular}[c]{@{}c@{}}$N_{2}$=4, C=128\\ r=4, M=4, $D_{r}$=3\\ G=2, K=7, $B_{r}$=3\end{tabular}}} 

& \multirow{3}{*}{\begin{tabular}[c]{@{}c@{}}$N_{2}$=4, C=192\\ r=4, M=6, $D_{r}$=3\\ G=2, K=7, $B_{r}$=3\end{tabular}}  \\
\multicolumn{1}{c|}{}                                                                                
& \multicolumn{1}{c|}{}                                                                               
& \multicolumn{1}{c|}{}                                                                                       
&                                                                                                  
\\
\multicolumn{1}{c|}{}                                                                                
& \multicolumn{1}{c|}{}                                                                                                    
& \multicolumn{1}{c|}{}                                                                                                    
&                                                                                                     
\\ \hline
\multicolumn{1}{c|}{\multirow{3}{*}{\begin{tabular}[c]{@{}c@{}}Stage 3\\ 14$\times$14\\ \end{tabular}}} 
& \multicolumn{1}{c|}{\multirow{3}{*}{\begin{tabular}[c]{@{}c@{}}$N_{3}$=8, C=256\\ r=2, M=8, $D_{r}$=3\\ G=4, K=5, $B_{r}$=3\end{tabular}}}  
& \multicolumn{1}{c|}{\multirow{3}{*}{\begin{tabular}[c]{@{}c@{}}$N_{3}$=18, C=256\\ r=2, M=8, $D_{r}$=3\\ G=4, K=5, $B_{r}$=3\end{tabular}}}  
& \multirow{3}{*}{\begin{tabular}[c]{@{}c@{}}$N_{3}$=18, C=384\\ r=2, M=12, $D_{r}$=3\\ G=4, K=5, $B_{r}$=3\end{tabular}} \\
\multicolumn{1}{c|}{}                                                                                
& \multicolumn{1}{c|}{}                                                                                                    
& \multicolumn{1}{c|}{}                                                                                                    
&                                                                                                     
\\
\multicolumn{1}{c|}{}                                                                                
& \multicolumn{1}{c|}{}                                                                                                    
& \multicolumn{1}{c|}{}                                                                                                    
&                                                                                                     
\\ \hline
\multicolumn{1}{c|}{\multirow{3}{*}{\begin{tabular}[c]{@{}c@{}}Stage 4\\ 7$\times$7\\ \end{tabular}}}   
& \multicolumn{1}{c|}{\multirow{3}{*}{\begin{tabular}[c]{@{}c@{}}$N_{4}$=2, C=512\\ r=1, M=16, $D_{r}$=3\\ G=8, K=3, $B_{r}$=3\end{tabular}}} 
& \multicolumn{1}{c|}{\multirow{3}{*}{\begin{tabular}[c]{@{}c@{}}$N_{4}$=6, C=512\\ r=1, M=16, $D_{r}$=1\\ G=8, K=3, $B_{r}$=2\end{tabular}}} 
& \multirow{3}{*}{\begin{tabular}[c]{@{}c@{}}$N_{4}$=4, C=768\\ r=1, M=24, $D_{r}$=3\\ G=8, K=3, $B_{r}$=3\end{tabular}} 
\\
\multicolumn{1}{c|}{}                                                                                
& \multicolumn{1}{c|}{}                                                                                                    
& \multicolumn{1}{c|}{}                                                                                                    
&                                                                                                     
\\
\multicolumn{1}{c|}{}                                                                                
& \multicolumn{1}{c|}{}                                                                                                    
& \multicolumn{1}{c|}{}                                                                                                    
&                                                                                                     
\\ \hline
\end{tabular}
\caption{DeBiFormer model architecture specifications. $N_{i}$: Number of blocks at stage $i$. $C$: Base channels in each block. $r$: Downsample ratio of deformed points. $M$: Number of attention heads in DBRMHA. $G$: Number of offset groups in DBRMHA. $D_{r}$: Deformable level MLP expansion ratio. $B_{r}$: Bi level MLP expansion ratio. $K$: Kernel size of offset module.}
\label{table-architecture}
%\vspace{-4mm}
\end{table}
%%%%%%%%%%%%%%%%%%%%%%%%%%%%%%%%%%%%%%%%%%%%%%%%%%%%%%%%%%%%%%%%%%%%%%%%%%%%%%%%%%%%%%%%%%%%%%%%%%%%%%%%%%%%%%%%%%%%%%
%%%%%%%%%%%%%%%%%%%%%%%%%%%%%%%%%%%%%%%%%%%%%%%%%%%%%%%%%%%%%%%%%%%%%%%%%%%%%%%%%%%%%%%%%%%%%%%%%%%%%%%%%%%%%%%%%%%%%%
Leveraging DBRA as a fundamental building block, we introduce a novel vision transformer called DeBiFormer. As depicted in Figure\ref{debi-stage}, we adhere to the recent state-of-the-art Vision Transformers \cite{CSWIN,SWIN,BIFORMER,DEF}, using a four-stage pyramid structure. 
In stage $i$, we utilize an overlapped patch embedding in the first stage and a patch merging module \cite{UNIFORMER,SSFORMER} in the second to fourth stages. This is done to decrease the input spatial resolution while increasing the number of channels. Subsequently, $N_{i}$ consecutive DeBiFormer blocks are used to transform the features. Within each DeBiFormer block, we adhere to recent methodologies \cite{UNIFORMER,MAXVIT,BIFORMER} by using a $3\times 3$ depthwise convolution at the outset. This is done to implicitly encode relative position information. Following that, we sequentially use a DBRA module with a 2-ConvFFN module with an expansion ratio $e$ for cross-location relation modeling and per-location embedding, respectively. 
DeBiFormer is instantiated in three distinct model sizes, achieved by scaling the network width and depth as outlined in Table \ref{table-architecture}. 
Each attention head comprises 32 channels, and we use a bi-level ConvFFN and deformable-level ConvFFN with an MLP expansion ratio of $e=3$. For the BRA, we use $topk = 1, 4, 16, S^{2}$, and for the DBRA, we use $topk = 4, 8, 16, S^{2}$ for the four stages. Moreover, we set the region partition factor $S$ to specific values: $S = 7$ for classification, $S = 8$ for semantic segmentation, and $S = 20$ for object detection tasks.
%%%%%%%%%%%%%%%%%%%%%%%%%%%%%%%%%%%%%%%%%%%%%%%%%%%%%%%%%%%%%%%%%%%%%%%%%%%%%%%%%%%%%
\section{Experiments}
\label{sec:formatting}
%%%%%%%%%%%%%%%%%%%%%%%%%%%%%%%%%%%%%%%%%%%%%%%%%%%%%%%%%%%%%%%%%%%%%%%%%%%%%%%%%%%%%%%%%%%%%%%%%%%%%%%%%%%%%%%%%%%%%%
\begin{table}[t]
  \begin{minipage}[t]{.47\linewidth}
    \begin{center}
    \small
      \begin{tabular}{l|ccc}
\hline
Model                   & \begin{tabular}[c]{@{}c@{}}FLOPs\\ (G)\end{tabular} & \begin{tabular}[c]{@{}c@{}}Params\\ (M)\end{tabular} & \begin{tabular}[c]{@{}c@{}}Top-1\\ (\%)\end{tabular} \\ \hline
ResNet-18 \cite{RN182}           & 1.8    & 11.7      & 69.8       \\
PVTv2-b1 \cite{LINFORMER}       & 2.1    & 13.1      & 78.7       \\
Shunted-T \cite{SSFORMER}       & 2.1    & 11.5      & 79.8       \\
QuadTree-B-b1 \cite{QuadTree}   & 2.3    & 13.6      & 80.0       \\
BiFormer-T \cite{BIFORMER}      & 2.2    & 13.1      & 81.4       \\
Conv2Former-N \cite{C2F}        & 2.2    & 15.0      & 81.5       \\
\rowcolor[HTML]{96FFFB}
DeBiFormer-T                   & 2.6    & 21.4         & \textbf{81.9}   \\ \hline
%Twins-PCPVT-B \cite{TWINS}      & 6.4    & 44        & 82.7      \\
PVTv2-B3 \cite{PVT}             & 6.9    & 45        & 81.2        \\
Swin-T \cite{SWIN}              & 4.5    & 29        & 81.3        \\
CSWin-T \cite{CSWIN}            & 4.5    & 23        & 82.7        \\
DAT-T \cite{DEF}                & 4.6    & 29        & 82.0        \\
CrossFormer-S \cite{CROSS}      & 5.3    & 31        & 82.5        \\
RegionViT-S+ \cite{RegionViT}   & 5.7    & 31        & 83.3        \\
QuadTree-B-b3 \cite{QuadTree}   & 7.8    & 46        & 83.7       \\
MaxViT-T \cite{MAXVIT}           & 5.6    & 31        & 83.6        \\
InternImage-T \cite{InternImage} & 5.0    & 30        & 83.5        \\
MixFormer-B4 \cite{MixFormer}    & 3.6    & 35        & 83.0       \\
BiFormer-S   \cite{BIFORMER}     & 4.5    & 26        & 83.8        \\
UniRepLKNet-T \cite{UniRepLKNet} & 4.9    & 31        & 83.2       \\
\rowcolor[HTML]{96FFFB}
DeBiFormer-S                    & 5.4       & 44          & \textbf{83.9}      \\ \hline
\end{tabular}
    \end{center}
    %\caption{Tiny, Small Models}
    %\label{IMAGENET}
  \end{minipage}
  \hfill
  \begin{minipage}[t]{.47\linewidth}
    \begin{center}
    \small
      \begin{tabular}{l|ccc}
\hline
Model                   & \begin{tabular}[c]{@{}c@{}}FLOPs\\ (G)\end{tabular} & \begin{tabular}[c]{@{}c@{}}Params\\ (M)\end{tabular} & \begin{tabular}[c]{@{}c@{}}Top-1\\ (\%)\end{tabular} \\ \hline
ConvNeXt-B \cite{Liu_2022_CVPR}    & 15.4   & 89        & 83.8       \\
SLaK-S \cite{SLaK-S}            & 9.8    & 55        & 83.8       \\
%HorNet-B \cite{HorNet}          & 15.5   & 88        & 84.3       \\
Twins-SVT-L \cite{TWINS}        & 14.8   & 99        & 83.7       \\
PVTv2-B5 \cite{PVT}             & 11.8   & 82        & 83.8       \\
Swin-B \cite{SWIN}              & 15.4   & 88        & 83.5       \\
Focal-B \cite{FOCAL}            & 16.4   & 90        & 84.0       \\
CSWin-B \cite{CSWIN}            & 15.0   & 78        & 84.2       \\
Shunted-B \cite{SSFORMER}       & 8.1    & 39        & 84.0       \\
UniFormer-B \cite{UNIFORMER}    & 8.3    & 50.0      & 83.9      \\
ScalableViT-B \cite{ScalableViT}& 8.6    & 81        & 84.1       \\
Slide-Swin-B \cite{SLIDE}       & 15.5   & 89.0      & 84.2       \\
DAT-S \cite{DEF}                & 9.0    & 50        & 83.7        \\
QuadTree-B-b4\cite{QuadTree}    & 11.5    & 64       & 84.1        \\
CrossFormer-L \cite{CROSS}      & 16.1   & 92        & 84.0       \\
RegionViT-B+ \cite{RegionViT}   & 13.6   & 74        & 83.8       \\
InternImage-S \cite{InternImage} & 8.0    & 50       & 84.2       \\
MixFormer-B6 \cite{MixFormer}    & 12.7   & 119      & 83.8       \\
BiFormer-B  \cite{BIFORMER}      & 9.8    & 57       & 84.3       \\
UniRepLKNet-S \cite{UniRepLKNet} & 9.1   & 56        & 83.9       \\
\rowcolor[HTML]{96FFFB}
DeBiFormer-B                     & 11.8      & 77     & \textbf{84.4}       \\ 
\hline
\end{tabular}
    \end{center}
    %\caption{Base Models}
  \end{minipage}
\caption{Evaluating and comparing different backbones on ImageNet-1K on images with resolution of $224 \times 224$.}
\label{FIGIMAGENET}
%\vspace{-4mm}
\end{table}
We experimentally evaluated the effectiveness of our proposed DeBiFormer on various mainstream computer vision tasks, including image classification (Section \ref{SS41}), semantic segmentation (Section \ref{S43}) and object detection and instance segmentation (Section \ref{S42}). In our approach, we commence training from scratch on ImageNet-1K \cite{IMAGENET} for image classification. Subsequently, we fine-tune the pre-trained backbones on ADE20K \cite{ADE20K} for semantic segmentation and on COCO \cite{COCO} for object detection and instance segmentation. Furthermore, we perform an ablation study to confirm the efficacy of the proposed Deformable Bi-level Routing Attention and top-k choices of DeBiFormer (Section \ref{S44}). Finally, in order to validate that the recognition ability and interpretability of our DeBiFormer, we visualize the attention map (Section \ref{S45}).

\subsection{Image classification on ImageNet-1K}\label{SS41}
\textbf{Settings.}
We conducted image classification experiments on the ImageNet-1K \cite{IMAGENET} dataset, following the experimental settings of DeiT \cite{DEIT} for a fair comparison. Specifically, each model was trained for 300 epochs on 8 V100 GPUs with an input size of $224 \times 224$. We used AdamW as the optimizer with a weight decay of $0.05$ and used a cosine decay learning rate schedule with an initial learning rate of $0.001$, while the first five epochs were used for linear warm-up. The batch size was set to 1024. To avoid overfitting, we used regularization techniques including RandAugment \cite{RAND} (rand-m9-mstd0.5-inc1), MixUp \cite{MIXUP} (prob = 0.8), CutMix \cite{CUTMIX} (prob = 1.0), Random Erasing (prob = 0.25), and increasing stochastic depth \cite{ER} (prob = 0.1/0.2/0.4 for DeBiFormer-T/S/B, respectively). 
\\
\textbf{Results.} 
We report our results in Table \ref{FIGIMAGENET} showing the top-1 accuracy with similar computational complexities. Our DeBiFormer outperformed the Swin Transformer \cite{SWIN}, PVT \cite{PVT}, DeiT \cite{DEIT}, DAT\cite{DEF}, and Biformer \cite{BIFORMER} in all three scales. Without inserting convolutions in Transformer blocks or using overlapped convolutions in patch embeddings, DeBiFormer achieved gains of 0.5pt, 0.1pt and 0.1pt over BiFormer \cite{BIFORMER} counterparts.  
%-------------------------------------------------------------------------
\subsection{Semantic segmentation on ADE20K}\label{S43}
%%%%%%%%%%%%%%%%%%%%%%%%%%%%%%%%%%%%%%%%%%%%%%%%%%%%%%%%%%%%%%%%%%%%%%%%%%%%%%%%%%%
\begin{table}[t]
\centering
\begin{tabular}{c|c|c}
\hline
\multirow{2}{*}{Backbone} & Semantic-FPN         & \multicolumn{1}{c}{UperNet} \\
                          & mIoU(\%)      & mIoU(\%)       \\ \hline
Swin-T \cite{SWIN}        & 41.5          & 44.5                 \\
DAT-T \cite{DEF}          & 42.6          & 45.5                \\
CSWin-T \cite{CSWIN}      & 48.2          & 49.3                 \\

RegionViT-S+ \cite{RegionViT}     & 45.3         & -                \\
InternImage-T \cite{InternImage}  & -            & 47.9           \\

CrossFormer-S \cite{CROSS}             & 46.0          & 47.6                 \\
Uniformer-S \cite{UNIFORMER}           & 46.6          & 47.6                \\
Shunted-S \cite{SSFORMER}              & 48.2          & 48.9                 \\
BiFormer-S \cite{BIFORMER}             & 48.9          & 49.8                  \\
%%%%%%%%%%%%%%%%%%%%%%%%%%%%%%%%%%%%%%%%%%%%%%%%%%%%%%%%%%%%%%%%%%%%%%%%%%%%%%%%%%%%%%
\rowcolor[HTML]{96FFFB}
DeBiFormer-S              & \textbf{49.2} & \textbf{50.0}           \\ \hline
Swin-S \cite{SWIN}                  & -             & 47.6                  \\
DAT-S \cite{DEF}                    & 46.1          & 48.3                  \\
CSWin-S \cite{CSWIN}                & 49.2          & 50.4                  \\

RegionViT-B+ \cite{RegionViT}     & 47.5         & -                 \\
InternImage-S \cite{InternImage}  & -            & 50.1            \\

CrossFormer-B \cite{CROSS}          & 47.7          & 49.7                  \\
Uniformer-B \cite{UNIFORMER}        & 48.0          & 50.0                 \\
BiFormer-B \cite{BIFORMER}          & 49.9          & 51.0                  \\
\rowcolor[HTML]{96FFFB}
DeBiFormer-B              & \textbf{50.6} & \textbf{51.4}            \\ \hline
\end{tabular}
\caption{Evaluating DeBiFormer on semantic segmentation with two segmentation heads (Semantic FPN and UpperNet) on ADE20K dataset.}
\label{ADE20K}
%\vspace{-4mm}
\end{table}
% -------------------------------------------------------------------------
%\textbf{Settings.} Following existing works, we conduct our semantic segmentation experiments on the ADE20K \cite{ADE20K} dataset based on MMSegmentation. We do comparisons under both Semantic FPN \cite{FPN} and UperNet \cite{UPERNET} frameworks. In both cases, the backbone is initialized with ImageNet-1K pretrained weights, and other layers use random initialization. Models are optimized with the AdamW optimizer and the batch size is set as 32. For a fair comparison, our Semantic FPN experiments use the same setting as PVT \cite{PVT} to train the model $80k$ steps. Our Upernet experiments use the same setting as Swin Transformer \cite{SWIN} to train the model 160k iterations.
\textbf{Settings.}
The same as existing works, we used our DeBiFormer on SemanticFPN \cite{FPN} and UperNet \cite{UPERNET}. In both cases, the backbone was initialized with ImageNet-1K pretrained weights. The optimizer was AdamW \cite{ADAM}, and the batch size was 32. For a fair comparison, we followed the same setting as PVT \cite{PVT} to train the model with 80k steps and Swin Transformer \cite{SWIN} to train the model with 160k steps.
\\
\textbf{Results.} 
Table \ref{ADE20K} shows the results of the two different frameworks. It shows that with the Semantic FPN framework, our DeBiFormer-S/B achieved 49.2/50.6 mIoU, respectively, improving BiFormer by 0.3pt./0.7pt. A similar performance gain for the UperNet framework was also observed. By utilizing the DBRA module, our DeBiFormer could caputure the most semantic key-value pairs, which makes the attention selection more reasonable and achieve higher performance on downstream semantic tasks.
%-------------------------------------------------------------------------
%-------------------------------------------------------------------------
\begin{table}[t]
\centering
\scalebox{0.9}{
\begin{tabular}{l|cccccccccccc}
\hline
\multirow{2}{*}{Backbone} 
& \multicolumn{6}{c|}{RetinaNet 1× schedule}                    
& \multicolumn{6}{c}{Mask R-CNN 1× schedule}                                      \\
& $mAP$  
& $AP_{50}$ 
& $AP_{75}$ 
& $AP_{S}$  
& $AP_{M}$  
&\multicolumn{1}{c|}{$AP_{L}$}  
& $mAP^{b}$ 
& $AP^{b}_{50}$
& $AP^{b}_{75}$
& $mAP^{m}$  
& $AP^{m}_{50}$
& $AP^{m}_{75}$
\\ 
\hline
Swin-T \cite{SWIN}            & 41.5 & 62.1 & 44.2 & 25.1 & 44.9 & \multicolumn{1}{c|}{55.5} & 42.2 & 64.6 & 46.2 & 39.1  & 61.6 & 42.0  \\

DAT-T \cite{DEF}              & 42.8 & 64.4 & 45.2 & 28.0 & 45.8 & \multicolumn{1}{c|}{57.8} & 44.4 & 67.6 & 48.5 & 40.4  & 64.2 & 43.1  \\

CSWin-T \cite{CSWIN}          & -    & -    &      & -    & -    & \multicolumn{1}{c|}{-}    & 46.7 & 68.6 & 51.3 & 42.2  & 65.6 & 45.4  \\

RegionViT-S+ \cite{RegionViT} & 43.9 & 65.5 & 47.3 & 28.5 & 47.3 & \multicolumn{1}{c|}{58.0} & 44.2 & 67.3 & 48.2 & 40.9  & 64.1 & 44.0  \\

MixFormer-B4 \cite{MixFormer} & -    & -    &      & -    & -    & \multicolumn{1}{c|}{-}    & 45.1 & 67.1 & 49.2 & 41.2  & 64.3 & 44.4  \\

CrossFormer-S \cite{CROSS}    & 44.4 & 55.3 & 38.6 & 19.3 & 40.0 & \multicolumn{1}{c|}{48.8} & 45.4 & 68.0 & 49.7 & 41.4  & 64.8 & 44.6  \\

QuadTree-B2 \cite{QuadTree}   & 46.2 & 67.2 & 49.5 & 29.0 & 50.1 & \multicolumn{1}{c|}{61.8} & -    & -    & -    & -     & -    & -     \\

InternImage-T \cite{InternImage} & -    & -    &      & -    & -    & \multicolumn{1}{c|}{-}    & 47.2 & 69.0 & 52.1 & 42.5  & 66.1 & 45.8  \\

Agent-Swin-T \cite{AgentAttention} & -    & -    &      & -    & -    & \multicolumn{1}{c|}{-}    & 44.6 & 67.5 & 48.7 & 40.7 & 64.4 & 43.4 \\

BiFormer-S \cite{BIFORMER}    & 45.9 & 66.9 & 49.4 & 30.2 & 49.6 & \multicolumn{1}{c|}{61.7} & 47.8 & 69.8 & 52.3 & 43.2  & 66.8 & 46.5  \\

\rowcolor[HTML]{96FFFB}
DeBiFormer-S                    & 45.6    & 66.6    & 48.9    & 28.7    & 49.3    & \multicolumn{1}{c|}{61.6}    & 47.5    & 69.7    & 52.1    & 42.5     & 66.2    & 45.7     \\ \hline

%%%%%%%%%%%%%%%%%%%%%%%%%%%%%%%%%%%%%%%%%%%%%%%%%%%%%%%%%%%%%%%%%%%%%%%%%%%%%%%%%%%%%%

Swin-S \cite{SWIN}            & 44.5 & 65.7 & 47.5 & 27.4 & 48.0 & \multicolumn{1}{c|}{59.9} & 44.8 & 66.6 & 48.9 & 40.9  & 63.4 & 44.2  \\
DAT-S \cite{DEF}              & 45.7 & 67.7 & 48.5 & 30.5 & 49.3 & \multicolumn{1}{c|}{61.3} & 47.1 & 69.9 & 51.5 & 42.5  & 66.7 & 45.4  \\
CSWin-S \cite{CSWIN}          & -    & -    & -    & -    & -    & \multicolumn{1}{c|}{-}    & 47.9 & 70.1 & 52.6 & 43.2  & 67.1 & 46.2  \\

RegionViT-B+ \cite{RegionViT} & 44.6 & 66.4 & 47.6 & 29.6 & 47.6 & \multicolumn{1}{c|}{59.0} & 45.4 & 68.4 & 49.6 & 41.6  & 65.2 & 44.8  \\

CrossFormer-B \cite{CROSS}    & 46.2 & 67.8 & 49.5 & 30.1 & 49.9 & \multicolumn{1}{c|}{61.8} & 47.2 & 69.9 & 51.8 & 42.7  & 66.6 & 46.2  \\
QuadTree-B3 \cite{QuadTree}  & 47.3 & 68.2 & 50.6 & 30.4 & 51.3 & \multicolumn{1}{c|}{62.9} & -    & -    & -    & -     & -    & -     \\

InternImage-S \cite{InternImage} & -    & -    &      & -    & -    & \multicolumn{1}{c|}{-}    & 47.8 & 69.8 & 52.8 & 43.3  & 67.1 & 46.7  \\

Agent-Swin-S \cite{AgentAttention} & -    & -    &      & -    & -    & \multicolumn{1}{c|}{-}    & 47.2 & 69.6 & 52.3 & 42.7 & 66.6 & 45.8 \\

BiFormer-B  \cite{BIFORMER}   & 47.1 & 68.5 & 50.4 & 31.3 & 50.8 & \multicolumn{1}{c|}{62.6} & 48.6 & 70.5 & 53.8 & 43.7  & 67.6 & 47.1  \\

\rowcolor[HTML]{96FFFB}
DeBiFormer-B                     & \textbf{47.1}    & 68.2    & 50.2    & 30.3    & \textbf{51.1}    & \multicolumn{1}{c|}{\textbf{63.0}}    & 48.5    & 70.2    & 53.3    & 43.2     & 67.2    & 46.4     \\ \hline
\end{tabular}
}
\label{COCO2017}
\vspace{+4mm}
\caption{Results on object detection (left group) and instance segmentation (right group) tasks, performed on COCO 2017 dataset.}
\end{table}
%-------------------------------------------------------------------------
\subsection{Object detection and instance segmentation}\label{S42}
\textbf{Settings.} 
We used our DeBiFormer as the backbone in the Mask RCNN \cite{RN18} and RetinaNet \cite{Retina} frameworks to evaluate the effectiveness of models for object detection and instance segmentation on COCO 2017 \cite{COCO}. The experiments were conducted with the MMDetection \cite{MMDET} toolbox. Before training on COCO, we initialized the backbone with weights pre-trained on ImageNet-1K and followed the same training strategies as BiFormer \cite{BIFORMER} to compare our methods fairly. Note that due to device limitations, we set mini batch size as \textbf{4} for these experiments, while in BiFormer this value is 16. For details on the specific settings of the experiment, please refer to the supplementary paper.
\\
\textbf{Results.} 
We list the results in Table \ref{COCO2017}. For object detection with RetinaNet, we report the mean average precision (mAP) and the average precision (AP) at different IoU thresholds (50\%, 75\%) for three object sizes (i.e., small, medium, and large (S/M/L)). From the results, we can see that while the overall performance of DeBiFormer was only comparable to some of the most competitive existing methods, the performance on large objects ($AP_L$) outperformed these methods although we use a limited resources. This may be because the DBRA allocates deformable points more reasonably. These points are not to focus only on small things, but to focus on important things in the image. Therefore the attention is not limited to a small area, which improves the detection accuracy of large objects. For instance segmentation with Mask R-CNN, we report the bounding box and mask the average precision ($AP_b$ and $AP_m$) at different IoU thresholds (50\%, 75\%). Note that our DeBiFormer still achieved great performance under the device limitation of mini batch size. We believe that we could achieve better results if the mini batch size could be the same to other methods since it has been proved on semantic segmentation tasks.

\subsection{Ablation study}\label{S44}
\textbf{Effectiveness of DBRA.}
We compared DBRA with several existing sparse attention mechanisms. Following CSWIN \cite{CSWIN}, we aligned macro architecture designs with Swin-T \cite{SWIN} for fair comparison. Specifically, we used $2, 2, 6, 2$ blocks for the four stages and non-overlapped patch embedding, and we set the initial patch embedding dimension to $C = 96$ and MLP expansion ratio to $e = 4$. The results are reported in Table \ref{EFFFFFFF}. Our Deformable Bi-level Routing Attention had significantly better performance than the existing sparse attention mechanisms, in terms of both image classification and semantic segmentation.
\begin{table}[t]
\centering
\begin{tabular}{c|c|c}
\hline
Sparse Attention & \begin{tabular}[c]{@{}c@{}}IN1K\\ Top1(\%)\end{tabular} & \begin{tabular}[c]{@{}c@{}}ADE20K\\ mIoU(\%)\end{tabular} \\ \hline
Shifted window \cite{SWIN} & 81.3 & 41.5 \\
Spatially sep \cite{TWINS} & 81.5 & 42.9 \\
Sequential axial \cite{SequentialAxia} & 81.5 & 39.8 \\
Criss-cross \cite{CROSS} & 81.7 & 43.0 \\
Cross-shaped window \cite{CSWIN} & 82.2 & 43.4 \\
Deformable \cite{DEF} & 82.0 & 42.6 \\
Block-grid \cite{MAXVIT} & 81.8 & 42.8 \\
Bi-level routing \cite{BIFORMER} & 82.7 & 44.8 \\
\rowcolor[HTML]{96FFFB}
Deformable bi-level routing & \textbf{82.9} & \textbf{48.0} 
\\ \hline
\end{tabular}
\caption{Ablation study on different attention mechanisms. All models follow the architecture design of the Swin-T model.}
\label{EFFFFFFF}
%\vspace{-6mm}
\end{table}
%-------------------------------------------------------------------------
\\
\textbf{Partition factor S.} 
Similar to BiFormer, we opted to use $S$ as a divisor of the training size to prevent padding. We used an image classification with a resolution of $224 = 7\times 32$, and we set $S = 7$ to ensure that the size of the feature maps was divided at each stage. This choice aligns with the strategy used for the Swin Transformer \cite{SWIN}, where a window size of 7 was used. 
\\
\textbf{Top-k Choices.}
%-------------------------------------------------------------------------
\begin{table}[t]
\centering
\begin{tabular}{c|c|c|c|c}
\hline
\#partition factor & $k$  & Bi-level Routing tokens to attend  & \begin{tabular}[c]{@{}c@{}}CoreML\\ Latency(ms)\end{tabular}  & \begin{tabular}[c]{@{}c@{}}Top-1 Acc\\ (\%)\end{tabular}   \\ \hline
7 & 1,4,16,49  & 64, 64, 64, 49             & 291      & 81.81 \\
7 & 2,8,32,49  & 128, 128, 128, 49          & 459      & 81.74 \\
7 & 4,8,16,32  & 256, 128, 64, 32          & \textbf{276}      & 81.47 \\
7 & 8,16,32,49 & 512, 256, 128, 49          & 498      & 81.60 \\ \hline
7 & 4,8,16,49  & 256, 128, 64, 49           & 382      & \textbf{81.90} \\ \hline
\end{tabular}
\caption{Ablation study on top-k selection.The four numbers in the k column represent the top-k values of the four stages and the same to the tokens to attend column. The CoreML Latency is conducted by MacBook Pro M1 using CPU and Nerual Engine. }
\label{Topktable}
%\vspace{-6mm}
\end{table}
%-------------------------------------------------------------------------
We systematically adjusted $k$ to ensure a reasonable number of tokens attended to deformable queries as the region size diminished in later stages. Exploring various combinations of $k$ is a viable option. In Table \ref{Topktable}, we present ablation results on IN-1K, following DeBiFormer-STL (“STL” denotes Swin-T Layout). A crucial observation from these experiments is that augmenting the number of tokens paid attention to the deformable queries had a detrimental effect on accuracy and latency, and increasing the number of tokens paid attention in stages 1 and 2 had an effect on accuracy.
\\
\textbf{Deformable Bi-level Routing Multi-Head Attention (DBRMHA) at different stages.} 
To evaluate the impact of design choices, we systematically replaced bi-level routing attention blocks with DBRMHA blocks across different stages, as shown in Table\ref{stageblock}. Initially, all stages used bi-level routing attention, similar to BiFormer-T \cite{BIFORMER}, achieving 81.3\% accuracy in image classification. Replacing just one block in the 4th stage with DBRMHA immediately boosted accuracy by +0.21. Replacing all blocks in the 4th stage added another +0.05. Further DBRMHA replacements in the 3rd stage continued to improve performance across tasks. While gains tapered off with earlier stage replacements, we settled on a final version—DeBiFormer—where all stages use Deformable Bi-level Routing Attention for simplicity.
% To investigate the impact of design choices, we systematically substituted bi-level routing attention blocks with DBRMHA blocks across different stages, as detailed in Table \ref{stageblock}. In the initial configuration (first row), all stages of the model consist of bi-level routing attention, akin to the block setup in BiFormer-T \cite{BIFORMER}. However, this version only achieved 81.3\% accuracy in image classification. We initiated the transition by replacing one bi-level routing attention block in the 4th stage with the proposed DBRMHA module, resulting in an immediate increase of +0.21 in accuracy. Subsequently, outfitting the entire 4th stage with DBRMHA yielded an additional improvement of +0.05. Continuing the process of DBRMHA replacements in the 3rd stage leads to further increments in performance across various tasks. These notable improvements underscore the efficacy of our proposed Deformable Bi-level Routing Attention. While the performance gains tend to saturate as DBRMHA is incorporated into earlier stages, we opt for the final version, where all four stages feature Deformable Bi-level Routing Attention, referred to as DeBiFormer, to maintain model simplicity.
\begin{table}[!t]
\centering
\begin{tabular}{cccc|cc|c}
\hline
\multicolumn{4}{c|}{Stage Configurations} & \multicolumn{2}{c|}{Backbone} & IN-1K \\
1                 & 2                 & 3                 & 4                 & FLOPs        & \#Param        & Acc.  \\ \hline
BB                & BB                & BB                & BB                & 2.2          & 13.1           & 81.37 \\
BB                & BB                & BB                & \textbf{B*}       & 2.4          & 15.8           & 81.58 \\
BB                & BB                & BB                & \textbf{**}       & 2.6          & 18.5           & 81.63 \\
BB                & BB                & \textbf{B*}       & \textbf{**}       & 2.7          & 21.2           & 81.84 \\
BB                & \textbf{B*}       & \textbf{B*}       & \textbf{**}       & 2.7          & 21.4           & 81.87 \\ 
\hline
\textbf{B*}       & \textbf{B*}       & \textbf{B*}       & \textbf{**}       & 2.6          & 21.4           & \textbf{81.90} \\ \hline
\end{tabular}
\caption{Configuration \textbf{B*}: this stage is constructed with successive bi-level routing attention and DBRMHA blocks, while BB and \textbf{**} denotes that stage uses the same blocks as bi-level routing attention and DBRMHA, respectively.}
\label{stageblock}
%\vspace{-8mm}
\end{table}
%-------------------------------------------------------------------------
\begin{figure}[t]
\begin{center}
\includegraphics[width=0.8\linewidth]{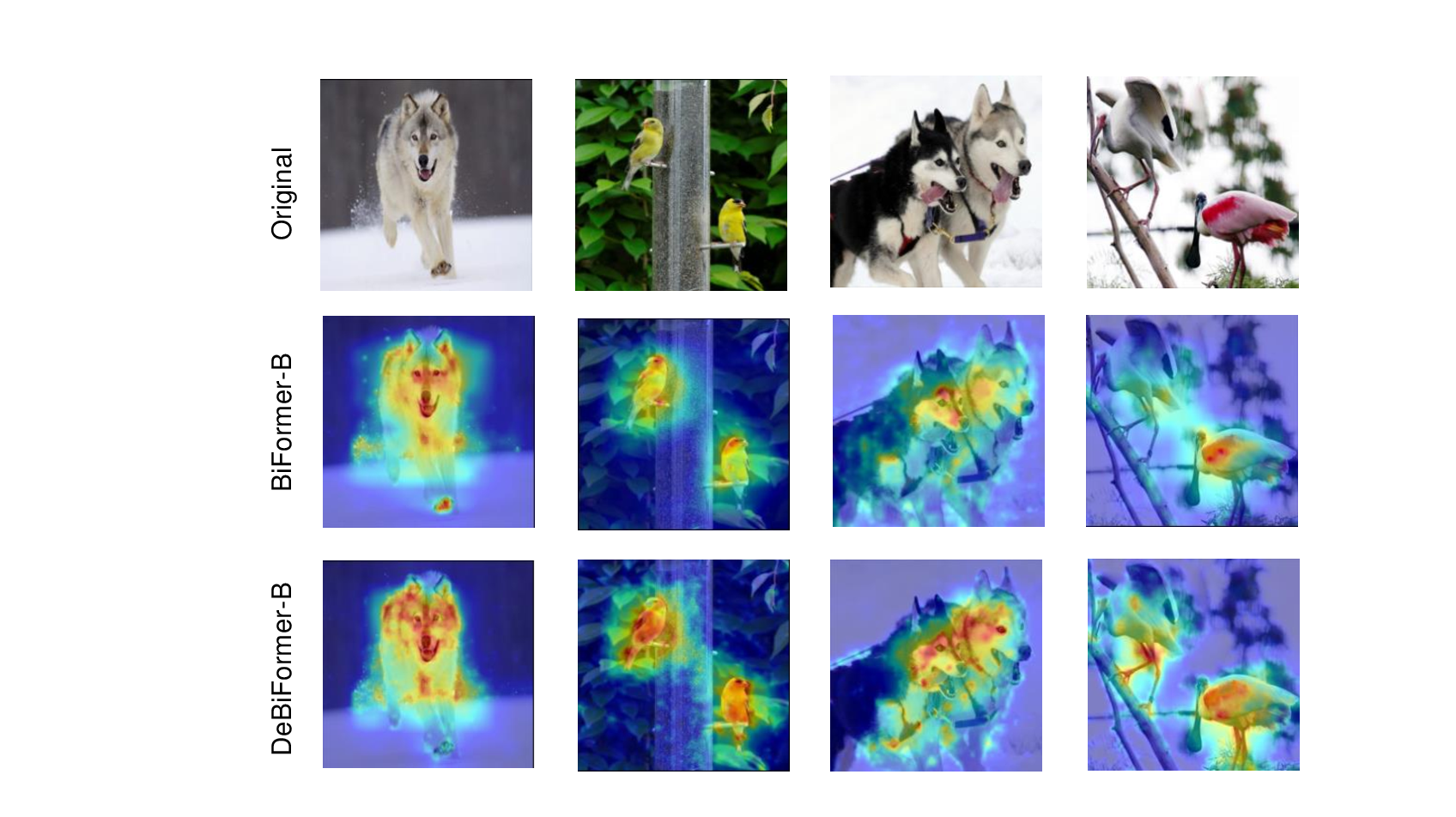}
\end{center}
   \caption{Grad-CAM Visualization of BiFormer-Base and DeBiFormer-Base. These images are sampled from the validation set of ImageNet-1K.}
\label{Grad-cam}
%\vspace{-3mm}
\end{figure}
%-------------------------------------------------------------------------
\section{Grad-CAM Visualization}\label{S45}
To further illustrate the ability of the proposed DeBiFormer to recognize the attention in important regions, we used Grad-CAM \cite{GRADCAM} to visualize the areas of greatest concern of BiFormer-Base and DeBiFormer-Base. 
As shown in Figure\ref{Grad-cam}, by utilizing DBRA module, our DeBiFormer-Base model performed better in target objects locating in which more regions have been focused on. In addition, our model scales down the attention in the unnecessary regions and pays more attention to the necessary regions. Depending on the attention of more necessary regions, our DeBiFormer model focused on semantic areas more continuously and completely, suggesting the stronger recognition ability of our model. Such ability yields better classification, and semantic segmentation performance compared with BiFormer-Base.
%As shown in Fig. \ref{Grad-cam}, our DeBiFormer-Base model performed better in locating the target objects and focusing on semantic areas more continuously and completely, suggesting the stronger recognition ability of our model. Such ability yields better classification, and semantic segmentation performance compared with BiFormer-Base.
%-------------------------------------------------------------------------
\section{Conclusion}
The paper introduces the Deformable Bi-level Routing Attention Transformer, a novel hierarchical vision transformer designed for both image classification and dense prediction tasks. Through Deformable Bi-level Routing Attention, our model optimizes query-key-value interactions while adaptively selecting semantically relevant regions. This leads to more efficient and meaningful attention. Extensive experiments show our model’s effectiveness compared to strong baselines. We hope this work offers insights into designing flexible and semantically aware attention mechanisms.
% This paper introduced the Deformable Bi-level Routing Attention Transformer, a novel hierarchical vision transformer designed to accommodate both image classification and dense prediction tasks. 
% By utilizing the Deformable Bi-level Routing Attention, we find a new way can not only optimize query focused by key and value pairs but also deformably choose the semantically relevant regions, which make the attention selection more reasonable.
% Extensive experiments demonstrate the effectiveness of our model over competitive baselines. We hope our work can inspire insights towards designing flexible and semantically relevant attention techniques.
\clearpage  % TODO REVIEW/FINAL: This \clearpage needs to be removed from both review and camera-ready versions.

\section{Supplementary Material}
\subsection{Offset groups}
Like \cite{DEF}, to foster diversity among deformed points, we adhere to a similar paradigm as in MHSA, where the channels are split into multiple heads to compute a variety of attentions. Hence, we partition the channels into $G$ groups to generate diverse offsets. The offset generation network shares weights for features from different groups.

\subsection{Deformable relative position bias}
Certainly, incorporating position information into attention mechanisms has proven beneficial for model performance. Various approaches, such as APE \cite{PVIT}, RPE \cite{SWIN}, CPE \cite{CPE}, LogCPB \cite{SWIN2}, and others have demonstrated improved results. The relative position embedding (RPE) introduced in the Swin Transformer specifically encodes the relative positions between every pair of query and key, thereby enhancing vanilla attention with spatial inductive bias \cite{SWIN}. The explicit modeling of relative locations is especially well-suited for attention heads at the deformable level. In this context, deformed keys can assume arbitrary continuous locations, as opposed to being confined to fixed discrete grids. 

In accordance with \cite{DEF}, the relative coordinate displacements are restricted to the range of $[-H, +H]$ and $[-W, +W]$ along the spatial dimension with a relative position bias (RPB), denoted as $\hat{B}$ with dimensions $(2H-1)\times (2W-1)$.

Then, the relative locations within the range of $[-1, +1]$ are sampled using bilinear interpolation $\varphi(\hat{B};R)$ with a parameterized bias. This is done by considering continuous relative displacements, ensuring coverage of all conceivable offset values.
 
\subsection{Computational complexity} 
Deformable bi-level routing attention (DBRA) incurs a comparable computation cost to its counterpart in the Swin Transformer. The computation of DBRA consists of two parts: token-to-token attention and offset \& sampling. Therefore, the computation of this part is:

\begin{align}
\begin{lgathered} 
FLOPs_{def} = FLOPs_{attn}+ FLOPs_{offset \& sampling}\\
= 2HWN_{s}C + 2HWC^{2} + 2N_{s}C^{2} + (k^{2}+6)N_{s}C
\end{lgathered}
\end{align}
where $N_{s} = HW/r^{2}$ is the number of sampled points, and $C$ is the token embedding dimension. The computation of bi-level routing multi-head attention consists of three parts: linear projection, region-to-region routing, and token-to-token attention. Hence, the computation of this part is:

\begin{align}
\begin{lgathered}  
  FLOPs_{bi} = FLOPs_{proj} + FLOPs_{routing} + FLOPs_{attn} \\
  = 2HWC^{2} + 2N_{s}C^{2} + 2(S^{2})^{2}C + 2HWk\frac{N_{s}}{S^{2}}C \\
  = 2HWC^{2} + 2N_{s}C^{2} + C\{2S^{4} + 2HWk\frac{N_{s}}{S^{2}}\} \\
  = 2HWC^{2} + 2N_{s}C^{2} + C\{2S^{4} + \frac{kHWN_{s}}{S^{2}} + \frac{kHWN_{s}}{S^{2}}\} \\
  \geq 2HWC^{2} + 2N_{s}C^{2} + 3C\{2S^{4} \cdot \frac{kHWN_{s}}{S^{2}} \cdot \frac{kHWN_{s}}{S^{2}}\}^{\frac{1}{3}} \\
  = 2HWC^{2} + 2N_{s}C^{2} + 3Ck^{\frac{2}{3}}\{2HWN_{s}\}^{\frac{2}{3}}
\end{lgathered}
\end{align}
where $k$ is the number of regions to attend to, and $S$ is the region partition factor. Last, the total computation of DBRA consists of two parts: deformable level multi-head attention and bi-level routing multi-head attention. The total amount of computations is therefore:

\begin{align}
\begin{lgathered} 
FLOPs = FLOPs_{bi}+ FLOPs_{def}\\
= 2HWN_{s}C + 2HWC^{2} + 2N_{s}C^{2} + (k^{2}+6)N_{s}C \\
+ 2HWC^{2} + 2N_{s}C^{2} + 3Ck^{\frac{2}{3}}\{2HWN_{s}\}^{\frac{2}{3}}\\
= 2HWN_{s}C + 4HWC^{2} + 4N_{s}C^{2} \\
+ (k^{2}+6)N_{s}C + 3Ck^{\frac{2}{3}}\{2HWN_{s}\}^{\frac{2}{3}}
\end{lgathered}
\end{align}
\\
In other words, DBRA achieves $O((HWN_{s})^{\frac{2}{3}})$ complexity. For example, the $3^{rd}$ stage of a hierarchical model with $224\times 224$ input for image classification usually has computation sizes of $H = W = 14\, S^{2} = 49\, N_{s} = 1\, C = 384$ and thus computational complexity for multi-head self-attention. Additionally, the complexity could be further reduced by enlarging the downsampling factor $r$ and scale with region partition factor $S$, making it friendly to tasks with much higher resolution inputs such as object detection and instance segmentation.

%%%%%%%%%%%%%%%%%%%%%%%%%%%%%%%%%%%%%%%%%%%%%%%%%%%%%%%%%%%%%%%%%%%%%%%%%%%%%%%%%%%
 \begin{table}[t]
 \centering
 \begin{tabular}{c|c|c}
 \hline
 Backbone & mIoU         & MS mIoU \\
 Swin-S \cite{SWIN}                  & 47.6          & 49.5                  \\
 DAT-S \cite{DEF}                    & 48.3          & 49.8                  \\
 CSWin-S \cite{CSWIN}                & 50.4          & 51.5                  \\

 InternImage-S \cite{InternImage}    & 50.1          & 50.9            \\

 CrossFormer-B \cite{CROSS}          & 49.7          & 50.6                  \\
 Uniformer-B \cite{UNIFORMER}        & 50.0          & 50.8                \\
 BiFormer-B \cite{BIFORMER}          & 51.0          & 51.7                  \\
 \rowcolor[HTML]{96FFFB}
 DeBiFormer-B              & \textbf{51.4} & \textbf{52.0}            \\ \hline
 \end{tabular}
 \caption{Evaluating DeBiFormer on semantic segmentation with two segmentation heads (UpperNet With MS IoU Scale) on ADE20K dataset.}
 \label{ADE20K}
 \end{table}
%%%%%%%%%%%%%%%%%%%%%%%%%%%%%%%%%%%%%%%%%%%%%%%%%%%%%%%%%%%%%%%%%%%%%%%%%%%%%%%%%%%%%
\subsection{Token to Attend} 
In Table \ref{Topktable}, we present the token to attend to the query and the token to attend to the deformable point. Compared with other methods, DeBiFormer has the fewest tokens to attend for each query but has a high performance in Imagenet1K, ADE20K(S-FPN head) and COCO(Retina head). 

\begin{table}[t]
\centering
\begin{tabular}{c|c|c|c|c}
\hline
\#Backbone &
\begin{tabular}[c]{@{}c@{}} Token to attend\\ for each query
\end{tabular}  & 
\begin{tabular}[c]{@{}c@{}}IN1K\\ Top1(\%)\end{tabular} &
\begin{tabular}[c]{@{}c@{}}ADE20K\\ mIoU(\%)\end{tabular} &
\begin{tabular}[c]{@{}c@{}}COCO\\ mAP(\%)\end{tabular}   \\ \hline
QuadTree-B3 \cite{QuadTree}   & 2048, 512, 128, 32  & 83.8 & 50.0 & \textbf{47.3}
\\
BiFormer-B \cite{BIFORMER}    & 64, 64, 64, 49 & 84.3 & 49.9 & 47.1  
\\ 
Debiformer-B  & 1, 1, 1, 1    & \textbf{84.4} & \textbf{50.6}  & 47.1  
\\
\hline     
\end{tabular}
\caption{Comparing different backbones Token to attend for each query in on ImageNet-1K, ADE20K, COCO.}
\label{Topktable}
\end{table}

\subsection{More visualization results}
\subsubsection{Effective Receptive Field Analysis}
\begin{figure}[!hb]
\begin{center}
\includegraphics[width=0.8\linewidth]{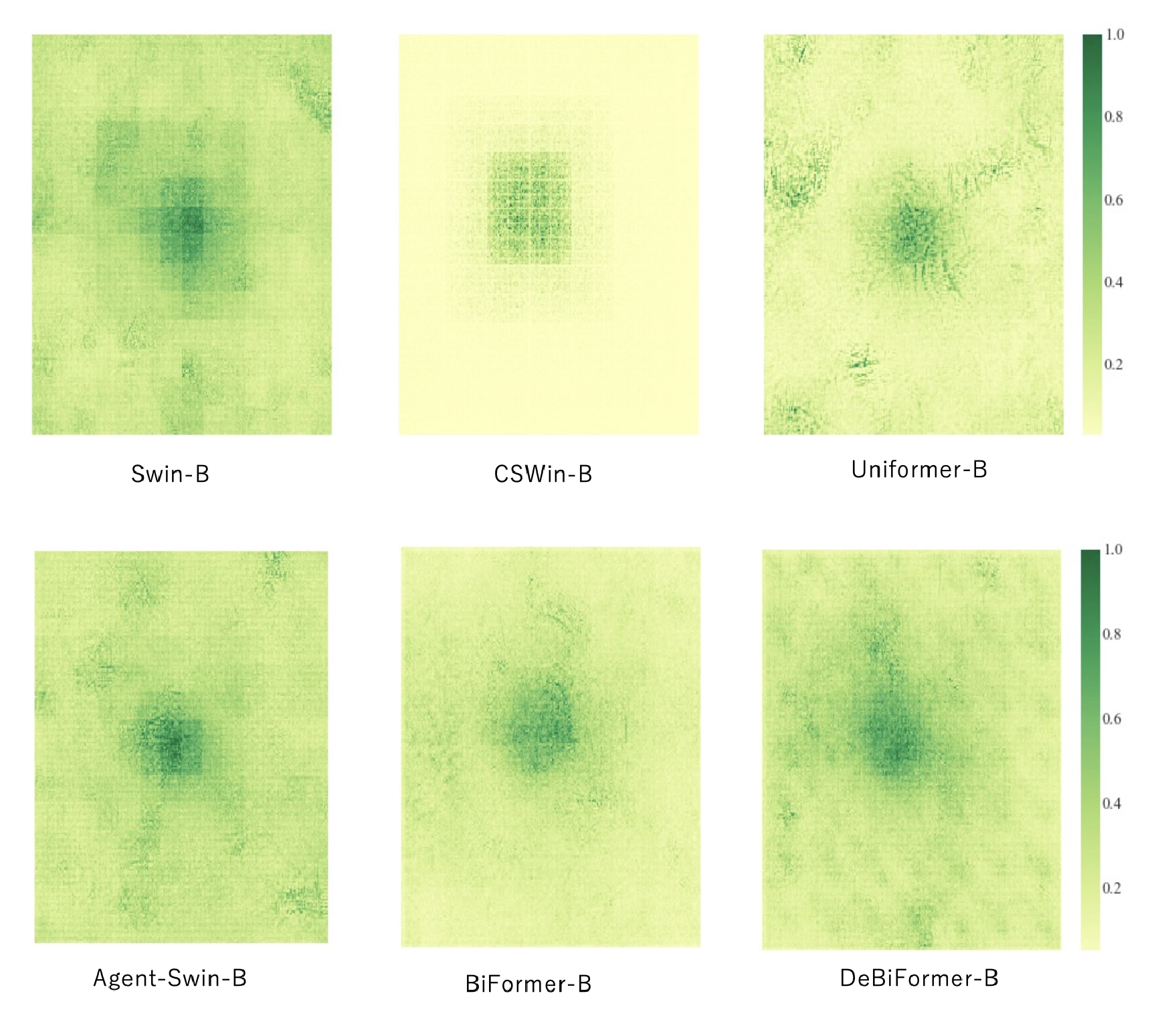}
\end{center}
   \caption{ERF visualization of models incorporating various local operators
and SOTA methods. The results are obtained by averaging over 100 images
(resized to 224×224) from ImageNet.}
\label{ERF}
\end{figure}

To assess the effective receptive fields (ERFs) \cite{ERF} of the central pixel with an input size of $224 \times 224$ across various models, we present a comparative analysis in Figure \ref{ERF}.
To demonstrate the powerful representation capacity of our DeBiFormer, we also compare the ERF of several SOTA methods with similar computational costs. As shown in Fig. \ref{ERF}, our DeBiFormer has the largest and most consistent ERF among these methods while maintaining strong local sensitivity, which is challenging to achieve.
%%%%%%%%%%%%%%%%%%%%%%%%%%%%%%%%%%%%%%%%%%%%%%%%%%%%%%%%%%%%%%%%%%%%%%%%
\subsubsection{Grad-CAM Analysis}
\begin{figure}[!h]
  \centering
  \begin{minipage}{0.95\linewidth}
      \includegraphics[width=\linewidth]{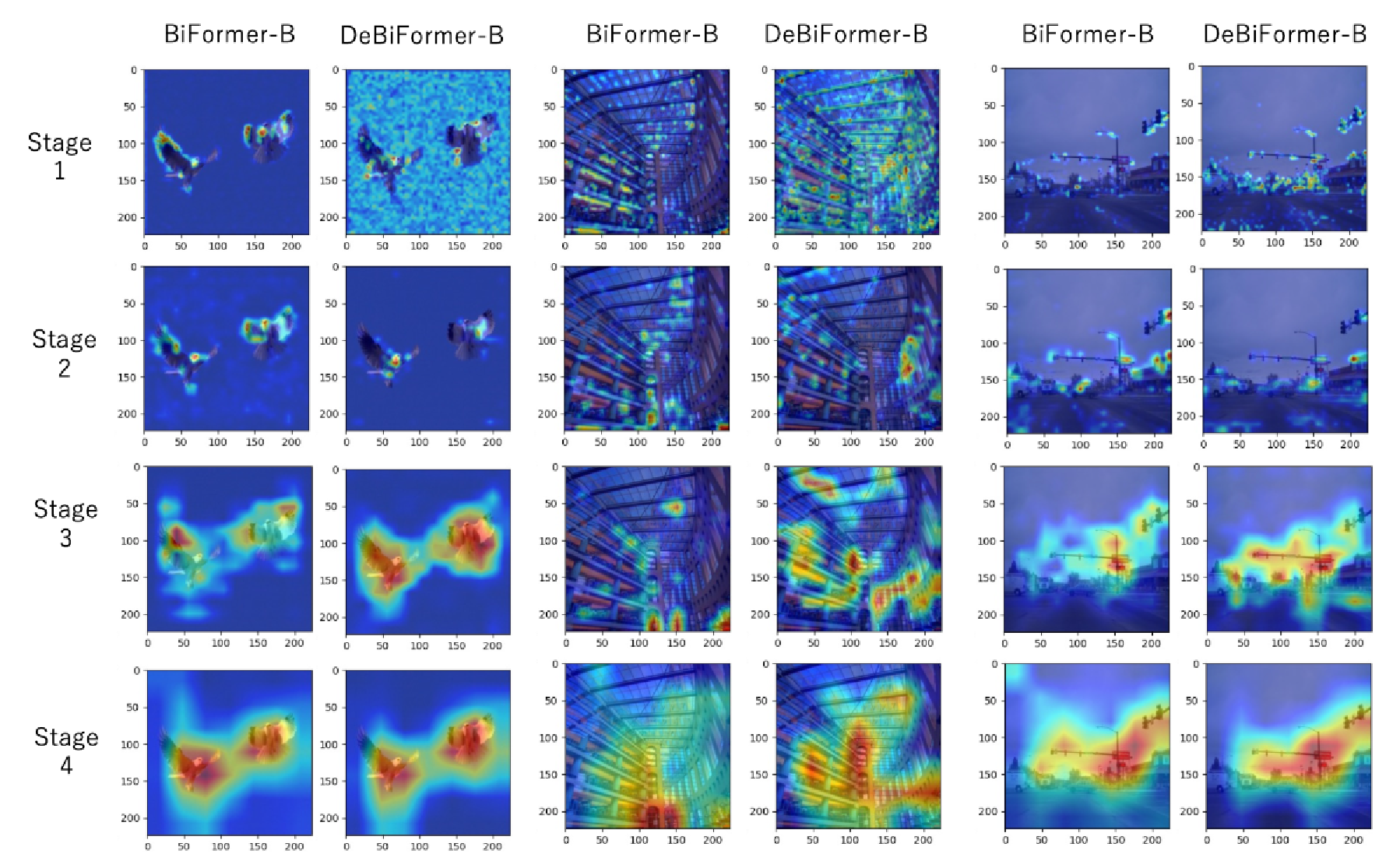}
  \end{minipage}\\
  \begin{minipage}{0.95\linewidth}
      \includegraphics[width=\linewidth]{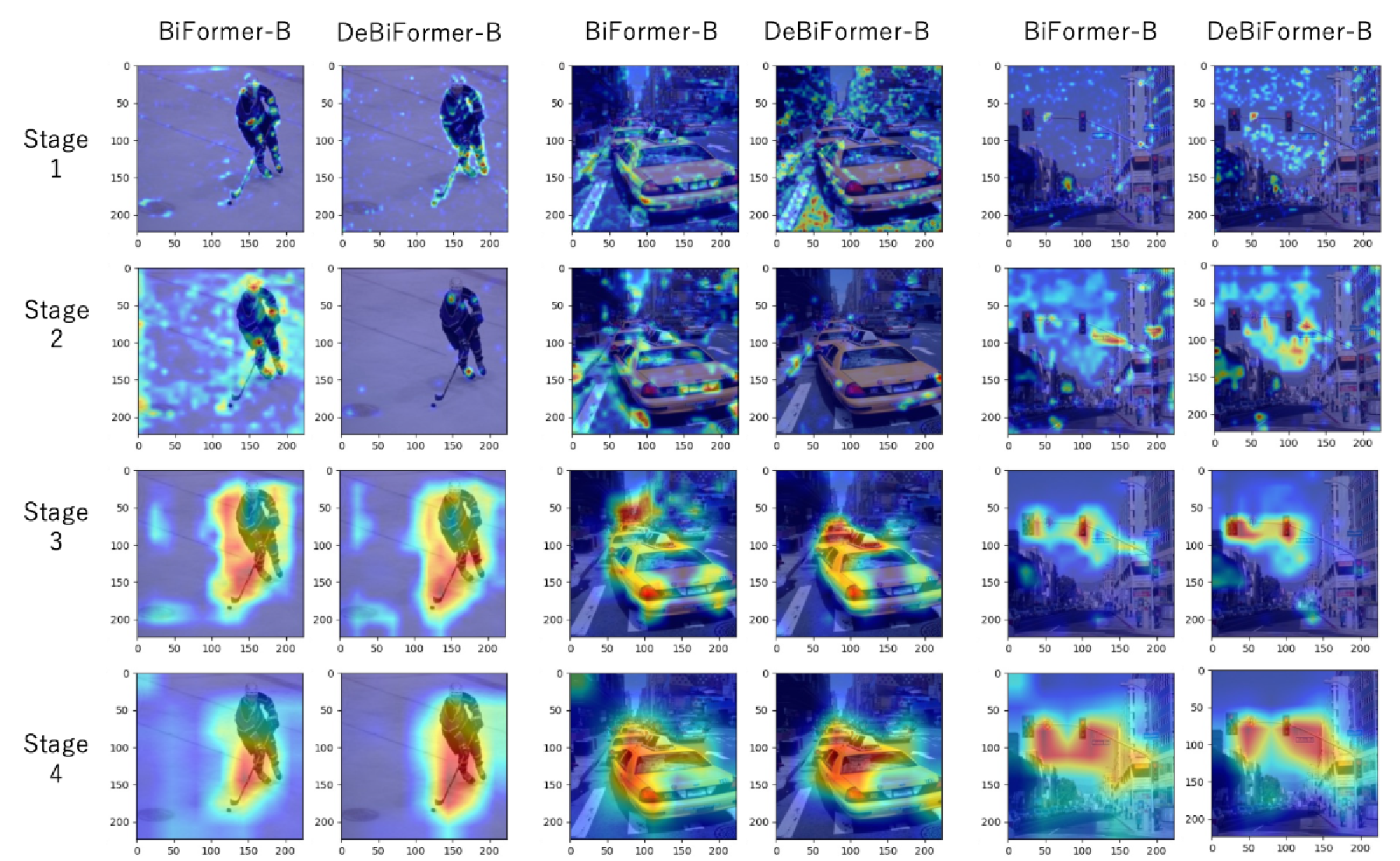}
  \end{minipage}
\caption{More Visualization of attention maps for different stages of DeBiFormer-Base. The attention maps show the focused regions has extracted large to small in each stage. Meanwhile the model fitted out the unessential regions to extract and enhance the important regions features in each stage.}
\label{atm}
\end{figure}

To further show how DBRA works, we demonstrate more visualization results in Figure \ref{atm}.
Thanks to the flexible key-value pairs selection, in most cases, our DeBiFormer focuses on important objects at an earlier stage. Meanwhile, due to the reasonable allocation of deformable points, it also focuses on different important areas earlier in multi-object scenarios. With the powerful DBRA module, our DeBiFormer has a larger heat map area in the last two stages, which represents a stronger ability of recognition.

%%%%%%%%%%%%%%%%%%%%%%%%%%%%%%%%%%%%%%%%%%%%%%%%%%%%%%%%%%%%%%%%%%%%%%%%
\subsection{Detailed Experimental Settings}
\begin{table}[tb]
\centering
\small
\begin{tabular}{c|ccc}
\hline
Settings              & DeBi-T & DeBi-S & DeBi-B \\ \hline
Input resolution      & 224    & 224    & 224    \\
Batch size            & 1024   & 512    & 512    \\
Optimizer             & AdamW  & AdamW  & AdamW  \\
Learning rate         & $1\times10^{-3}$ & $5\times10^{-4}$ & $5\times10^{-4}$ \\
LR schedule           & cosine & cosine & cosine \\
Weight decay          & $5\times10^{-2}$ & $5\times10^{-2}$ & $5\times10^{-2}$ \\
Warmup epochs         & 20     & 20     & 20     \\
Epochs                & 300    & 300    & 300    \\ \hline
Horizontal flip       &  \checkmark      &  \checkmark      &  \checkmark      \\
Random resize         &  \checkmark      &  \checkmark      &  \checkmark      \\
AutoAugment           &  \checkmark      &  \checkmark      &  \checkmark      \\
Mixup alpha           & 0.8    & 0.8    & 0.8    \\
Cutmix alpha          & 1.0    & 1.0    & 1.0    \\
Random erasing prob.  & 0.25   & 0.25   & 0.25   \\
Color jitter          & 0.4    & 0.4    & 0.4    \\ \hline
Label smoothing       & 0.1    & 0.1    & 0.1    \\
%Dropout               & ✗      & ✗      & ✗      \\
Droppath rate         & 0.     & 0.2    & 0.4    \\
%%Repeated augment      & ✗      & ✗      & ✗      \\
Grad. clipping        & 5.0    & 5.0    & 5.0    \\ \hline
\end{tabular}
\caption{Image Classification Training Settings}
\label{imagenet}
\end{table}
%%%%%%%%%%%%%%%%%%%%%%%%%%%%%%%%%%%%%%%%%%%%%%%%%%%%%%%%%%%%%%%%%%%%%%%%%%%%%%%%%%%%%
\subsubsection{Image classification on ImageNet-1K}
As introduced in the main paper, each model was trained for 300 epochs on 8 V100 GPU with an input size of $224 \times 224$. The experimental settings are strictly follow the DeiT\cite{DEIT} for a fair comparison. For more details, please refer to the Table \ref{imagenet} provided.

\begin{table}[t]
\centering
\small
\begin{tabular}{c|c}
\hline
Config & Value \\ \hline
Optimizer & AdamW \\
LR & 0.0002 \\
weight decay & 0.05 \\
batch size & \textbf{4} \\
LR schedule & steps:[8, 11]  \\
training & epochs 12 \\
scales & (800, 1333) \\
drop path & 0.2 (Small), 0.3 (Base) \\ \hline
\end{tabular}

\caption{ Object Detection and Instance Segmentation Training Settings}
\label{coco}
\end{table}

\subsubsection{Object Detection and Instance Segmentation}
When fine-tuning our DeBiFormer to object detection and instance segmentation on COCO \cite{COCO}, we have considered two common frameworks: Mask R-CNN \cite{RN18}, RetinaNet \cite{Retina}. For optimization, we adopt the AdamW optimizer with an initial learning rate of 0.0002 and a \textbf{mini batch size of 4} due to the limitation of devices. When training models of different sizes, we adjust the training settings according to the settings used in image classification. The detailed hyper-parameters used in training models are presented in Table \ref{coco}.

\subsubsection{Semantic Segmentation}
For ADE20K, we utilized the AdamW optimizer with
an initial learning rate of 0.00006, a weight decay of 0.01, and a mini batch size of 16 for all models trained for 160K iterations. In terms of testing, we reported the results using both single-scale (SS) and multi-scale (MS) testing in the main comparisons. For multi-scale testing, we experimented with resolutions ranging from 0.5 to 1.75 times that of the training resolution. To set the path drop rates in different models, we used the same hyper-parameters as those used for object detection and instance segmentation. Table \ref{ADE20K} shows the results of the Upernet frameworks with the single and Multi-Scale IoU.

\subsection{Limitation and Future Work}
In contrast to sparse attention with simple static patterns, we propose a novel attention method that consists of two components. First, we prune a region-level graph and gather key-value pairs for important regions, which are focused by highly flexible key-value pairs. Then, we apply token-to-token attention. While this method does not incur much computation as it operates at a top-k routed semantically relevant region level and deformable important regions, it inevitably involves additional parameter capacity transactions during linear projection. In our future endeavors, we plan to investigate efficient sparse attention mechanisms and enhance Vision Transformers with parameter capacity awareness.

\bibliographystyle{splncs04}
\bibliography{main}
\end{document}